\documentclass[11pt,a4paper]{article}
\usepackage{times,latexsym}
\usepackage{url}
\usepackage[T1]{fontenc}
\usepackage[acceptedWithA]{tacl2021v1}
\usepackage{amsmath}
\usepackage{amssymb}
\usepackage{xcolor}
\usepackage{booktabs}
\usepackage{multirow}
\usepackage{graphicx}
\usepackage{enumitem}
\usepackage{adjustbox}
\usepackage{makecell}
\usepackage{xurl}
\makeatletter
\renewcommand\@makefntext[1]{%
  \noindent\makebox[1em][l]{\@makefnmark}#1}
\makeatother
\usepackage{listings}
\newcommand{\dataname}{\texttt{InsightTrace}}
\newcommand{\modelname}{\texttt{InsightShield}}
\newcommand{\tasknameauthorship}{\texttt{{\small Textual Authorship}}}
\newcommand{\tasknameinsightprovenance}{\texttt{{\small Insight Provenance}}}
\newcommand{\llmparaphrase}{\texttt{{\small LLM-Polish}}}
\newcommand{\llmenrichbullets}{\texttt{{\small LLM-Expand}}}
\newcommand{\fullyhumanreview}{\texttt{{\small Human-Review}}}
\newcommand{\fullyllmreview}{\texttt{{\small LLM-Review}}}

\newcommand{\labelhumaninsight}{\texttt{{\small Human-Insight}}}
\newcommand{\labelhybrid}{\texttt{{\small Hybrid-Insight}}}
\newcommand{\labelaiinsight}{\texttt{{\small AI-Insight}}}
\newcommand{\labelhumanwritten}{\texttt{{\small Human-Written}}}
\newcommand{\labelaiwritten}{\texttt{{\small AI-Written}}}

\title{Who Wrote It Is Not Enough: Detecting Who Contributed the Insight}

\author{
  Zhuoyang Zou$^{1}$ \quad Abolfazl Ansari$^{1}$ \quad Jiaxi Yang$^{1}$ \quad Delvin Ce Zhang$^{2}$ \\
  \textbf{Qian Chen$^{1}$\thanks{Corresponding author.} \quad Dongwon Lee$^{1}$ \quad Wenpeng Yin$^{1}$} \\[2pt]
  $^{1}$Pennsylvania State University, USA \quad
  $^{2}$University of Sheffield, United Kingdom \\[2pt]
  \texttt{\{zhuoyangzou, wenpeng\}@psu.edu}
}
\date{}

\begin{document}
\maketitle

\begin{abstract}
As LLMs increasingly assist scientific writing and peer review, detecting \emph{who wrote the text} (\tasknameauthorship) is no longer sufficient: we need to determine \emph{who contributed the underlying insight}. We introduce \tasknameinsightprovenance, the task of identifying whether a review insight originates from a human, an LLM, or their hybrid contribution. We construct \dataname~from 4,057 scientific papers and 12,660 human reviews, simulating a spectrum of human--LLM collaboration---from paraphrasing human-written reviews, to expanding human-provided critiques, to generating reviews entirely from the paper---using GPT-4o, Gemini, and DeepSeek. We propose \modelname, a shortcut-unlearning  framework that adversarially removes linguistic and \tasknameauthorship~shortcuts while retaining signals predictive of \tasknameinsightprovenance. Our further analyses reveal what makes intellectual authorship identifiable: paper grounding and neighboring review context provide complementary provenance signals, while human, hybrid, and AI insights systematically differ in their information sources and failure modes.\footnote{Data and code will be released upon publication.}
\end{abstract}

\section{Introduction}

As LLMs increasingly participate in scientific writing and peer review \citep{DBLP01268,DBLPAlchokrSSLK24}, a fundamental attribution question is becoming urgent: \emph{who actually contributed the insight?} A human reviewer may independently identify a weakness and use an LLM only to polish the wording, or may delegate the substantive critique itself to an LLM. Both outputs can be AI-written, yet their intellectual provenance is fundamentally different. Attribution and accountability therefore require distinguishing \emph{who wrote the words} from \emph{who contributed the insight}\footnote{An insight is the substantive proposition expressed by a review sentence about the paper, such as an observation, evaluation, criticism, comparison, or recommendation, abstracted away from the particular wording used to express it.}.

Existing AI-text detection largely addresses the former---\tasknameauthorship---by identifying linguistic or statistical signatures of machine-generated text \citep{DBLP:conf/icml/Mitchell0KMF23,DBLP:conf/iclr/BaoZTY024,DBLP:conf/acl/SuWWZL25}. We instead introduce \tasknameinsightprovenance: determining whether the insight expressed in a review sentence originates from a \textit{human}, an \textit{LLM}, or their \textit{hybrid}\footnote{Human provides part of the underlying insight and the LLM develops additional substantive content.} contribution. To our knowledge, prior work has not directly studied this distinction at the level of scientific-review insights.

No existing dataset captures the provenance of insights contributed during scientific peer review. We construct \dataname~from 4,057 scientific papers and 12,660 human reviews, simulating progressively greater LLM involvement through four generation conditions: \fullyhumanreview, \llmparaphrase, \llmenrichbullets, and \fullyllmreview, using GPT-4o \citep{DBLP:journals/corr/abs-2410-21276}, Gemini-3.5-Flash-Lite \citep{DBLP:journals/corr/abs-2312-11805}, and DeepSeek-V3.1 \citep{DBLP:journals/corr/abs-2412-19437}. Importantly, above generation conditions do not determine provenance: we further annotate each review sentence according to whether its underlying insight is fully, partially, or not supported by the human-provided source. We also construct progressively debiased test sets that remove stylistic and rhetorical shortcuts, explicitly testing whether models detect insight provenance rather than surface authorship.

We find that this distinction matters empirically. Models achieve deceptively strong performance on raw data by exploiting linguistic and authorship shortcuts, but performance drops substantially (from $\sim$90\% to $\sim$60\% F1) once these cues are normalized. To address this, we design a shortcut-unlearning framework, \modelname. It first learns shortcut-rich representations, then freezes the resulting provenance and authorship detectors as adversaries and learns a new representation that reduces their predictions toward random guessing while preserving signals predictive of insight provenance. This explicitly suppresses the signals responsible for the original shortcut solution while preserving provenance-relevant information.

Most importantly, our analysis reveals \emph{why insight provenance remains detectable even after linguistic normalization}. AI-origin insights overwhelmingly stay close to information already available: $44.5\%$ are field-generic and $38.6\%$ are explicit in the paper. Human insights, in contrast, much more often introduce external knowledge ($28.0\%$ vs.\ $2.7\%$) or independent reviewer judgment ($28.9\%$ vs.\ $11.1\%$). Surprisingly, paper-inferable reasoning is similarly rare across all classes. The distinction is therefore not simply that humans ``reason harder'' over the same paper, but that they more often bring \emph{information and judgment from beyond the paper}. This suggests that intellectual authorship leaves a signal of \emph{provenance distance}: style can be rewritten, but where an idea came from is harder to erase.

Our contributions are fourfold: 

i) We introduce \tasknameinsightprovenance, shifting AI detection from \emph{who wrote the text} to \emph{who contributed the underlying insight}. 

ii) We construct \dataname, a sentence-level benchmark of human, hybrid, and AI-origin scientific-review insights across three LLM families, together with progressively debiased evaluation sets. 

iii) We propose \modelname, an adversarial representation-learning framework designed to suppress authorship and other shortcut signals while preserving insight provenance. 

iv) Through extensive analyses, we uncover what makes \tasknameinsightprovenance~detectable beyond writing style, revealing the roles of paper grounding and local review context, systematic differences among human, hybrid, and AI insights, and the information sources and failure modes underlying provenance detection.

\section{Related Work}

\vspace{-2mm}
\paragraph{\tasknameauthorship.}
A large body of work studies whether observed text was written by a human or generated by an LLM. Early and representative approaches formulate this as binary human--AI detection, using model likelihoods or statistical signatures, such as DetectGPT \citep{DBLP:conf/icml/Mitchell0KMF23}, Fast-DetectGPT \citep{DBLP:conf/iclr/BaoZTY024}, and Binoculars \citep{DBLP:conf/icml/HansSCKSGGG24}, or supervised classifiers trained on human and machine text, such as HC3 \citep{DBLP:journals/corr/abs-2301-07597} and CoCo \citep{DBLP:conf/emnlp/LiuZWPL023}. Subsequent work increasingly focuses on robustness and generalization: MAGE \citep{DBLP:conf/acl/LiLCBWWY0024} systematically studies cross-domain, cross-generator, and paraphrasing settings; RADAR \citep{DBLP:conf/nips/HuCH23} improves paraphrase robustness through adversarial training; and DeTeCtive \citep{DBLP:conf/nips/GuoHZZFHM24} learns fine-grained source representations to better generalize across generators and domains.
More recent work moves beyond document-level binary detection toward \emph{fine-grained and collaborative authorship attribution}. SeqXGPT \citep{DBLP:conf/emnlp/WangLRJZQ23} performs sentence-level attribution within documents containing both human- and AI-written text, while HACo-Det \citep{DBLP:conf/acl/SuWWZL25} further localizes human versus AI contributions at the word level. Recognizing that real writing may involve both humans and LLMs, MixSet \citep{DBLP:conf/naacl/ZhangGCHHSZLFWS24} studies AI-revised human text and human-revised AI text, while DETree \citep{DBLP:conf/nips/HeZCML25} and FAID \citep{DBLP:conf/eacl/TaVHLNNWNS26} model increasingly diverse human--AI collaboration and editing processes.

\vspace{-2mm}
\paragraph{Accountability and Integrity in AI-Assisted Peer Review.}
The growing use of LLMs in scientific peer review has raised concerns about the integrity, transparency, and accountability of the reviewing process \citep{leimeister2026chatgpt,van2026beyond}. In response, \citet{kumar-etal-2024-quis} develop peer-review-specific detectors for identifying ChatGPT-written reviews, while \citet{yu2026is} systematically benchmark AI-text detectors on large-scale human and LLM-generated peer reviews. However, the increasing prevalence of mixed human--AI reviewing complicates such detection. \citet{saha2026policies} show that existing detectors can mistake LLM-polished human reviews for AI-generated ones, making it difficult to enforce policies that permit linguistic assistance while prohibiting substantive delegation to LLMs. More broadly, recent discussions of scholarly accountability argue that disclosure of AI use alone is insufficient when it remains unclear who is responsible for the reasoning and judgments expressed in scholarly communication \citep{van2026beyond}. These concerns motivate a distinction beyond whether AI participated in \emph{writing} a review: an LLM may merely verbalize a critique independently identified by a human reviewer, or it may introduce a substantive scientific judgment of its own. Our work operationalizes this distinction through \tasknameinsightprovenance, attributing individual review insights to human, AI, or joint intellectual contribution rather than merely detecting AI-generated text.

\vspace{-3mm}
\paragraph{\tasknameinsightprovenance.} Very recently, RACE \citep{DBLP:conf/acl/LiSWYWC26}, published only a few months prior to our work, moves beyond identifying the final writer by separately modeling the \emph{creator} and \emph{editor} of a text, distinguishing, for example, human-created text polished by an LLM from LLM-generated text subsequently humanized. Its focus, however, remains on \emph{production provenance}: who established the textual and rhetorical foundation and who edited its surface realization. Our work differs from RACE in three key aspects: i) \textit{Creator $\neq$ insight contributor.} RACE models who initiates a text and who subsequently edits it, assuming that the creator establishes its substantive foundation. However, being the creator does not imply originating every insight. For example, a human may initiate a review by providing several critique bullets, while the LLM, when developing them into a full review, introduces an additional critique not present in the human input. The review is human-initiated, yet this new insight is AI-originated. ii) \textit{Document vs.\ proposition.} RACE assigns creator/editor roles to a document as a whole. Such document-level attribution is insufficient for our setting: even under the same human--LLM workflow, the LLM may preserve some human insights while extending others or introducing entirely new ones. Consequently, a single review can contain human-, hybrid-, and AI-originated insights, making the individual proposition the appropriate unit of provenance. iii) \textit{Intrinsic vs.\ relational evidence.} RACE infers production history from the observed text and its rhetorical structure. Our \tasknameinsightprovenance~is relational: determining where an insight originates requires relating the review sentence to its source paper and surrounding review context.


\section{Dataset \dataname~Construction}
\vspace{-2mm}
\paragraph{Paper and human-written review collection.} We scrape OpenReview for papers whose full submission PDF and complete, non-anonymised
official reviews are both public. The pool contains \textbf{4{,}057 papers} and
\textbf{13{,}600 reviews}, of which \textbf{12{,}660} have non-empty review bodies and
are retained. The corpus is dominated by ICLR 2017--2021 (ICLR~2019: 1{,}399 papers;
2018: 884; 2020: 884; 2017: 414; 2021: 140), with a small tail from NeurIPS~2021, MIDL
2018--2021, AKBC 2019--2021 and CoRL~2021. 
Submission PDFs are converted to text with PyMuPDF\footnote{\url{https://github.com/pymupdf/pymupdf}}.
The pre-2022 cutoff is deliberate: it predates the widespread availability of instruction-tuned LLMs, making it unlikely that the human half of the corpus is contaminated by LLM assistance. Those human-written reviews are marked as \fullyhumanreview.

\vspace{-2mm}
\paragraph{Generating reviews with different levels of LLM involvement.}
To capture this spectrum, we construct three review-generation settings with progressively increasing LLM involvement: \llmparaphrase, \llmenrichbullets, and \fullyllmreview. To further test whether an \tasknameinsightprovenance~detector generalizes beyond a particular LLM, we instantiate the AI assistant with three LLMs: GPT-4o, Gemini-3.5-Flash-Lite, and DeepSeek-V3.1.

\textbullet\enspace\textbf{\llmparaphrase.}
This setting models a common writing-assistance scenario in which the reviewer develops the substantive arguments but uses an LLM only to improve their presentation. Specifically, we ask the LLM to paraphrase the original \fullyhumanreview. Thus, although the surface realization is rewritten by an LLM, the underlying critiques and insights are intended to remain human-originated.

\textbullet\enspace\textbf{\llmenrichbullets.}
This setting models a collaborative workflow in which the human provides the
core critique points and the LLM develops them into a complete review. Since
\fullyhumanreview~contains full reviews rather than pre-writing notes, we  i) first ask an LLM to extract 3--5 original
weakness descriptions \emph{verbatim} from \fullyhumanreview, without
modification, paraphrasing, or addition (Appendix~\ref{app:prompt-extract}). We retain only extracted text that
can be exactly matched to the original human review, ensuring that all input
critiques are human-written, ii) then provide these extracted
weaknesses together with the paper to an LLM and asks it to develop them into
a complete review. This creates an intermediate setting in which part of the critique is
guaranteed to be human-originated, while the complete review may also contain
new insights introduced by the LLM as it interprets the paper and elaborates
the extracted human critiques.

\textbullet\enspace\textbf{\fullyllmreview.}
This setting represents the LLM-dominant endpoint of the spectrum, modeling cases in which a reviewer delegates the substantive review-generation process to an LLM rather than first supplying their own critiques. The LLM receives the paper but not the original human review and is instructed to generate a review from scratch.\footnote{To avoid tying this condition to a single prompting style, we sample uniformly
from a pool of 34 review requests spanning different formats, emphases, and evaluation rubrics, ranging from ``\textit{Review this paper}'' to ``\textit{Assess methodological rigor (baselines, ablations, stats, datasets) and list missing experiments}''; the full pool is listed in Table~\ref{tab:prompt-pool} (Appendix~\ref{sec:appendix-data})}. Because the human review is completely ignored, the substantive critiques produced in this setting are attributable to the LLM rather than inherited from the original reviewer.

\textbf{Question: Why not consider human-edited AI reviews?} 
Answer: While such reviews likely occur in practice, obtaining reliable data is challenging. Controlled collection would require experts to edit AI-generated reviews with span-level edit tracking, but this artificial setting may not reflect natural reviewing behavior because participants know their edits will be subsequently examined, potentially altering both their editing behavior and the quality of their revisions. Ecologically valid data would therefore ideally come from real conference or journal reviewing workflows, where such provenance is rarely observable.

\begin{table}[t]
\centering
\small
\setlength{\tabcolsep}{3pt}
\begin{tabular}{@{}llrrr@{}}
\toprule
\textbf{LLM} & \textbf{Involve.} & \textbf{Human} & \textbf{Hybrid} & \textbf{AI} \\
\midrule
\multirow{2}{*}{GPT-4o}
  & LLM-Polish & 90.6 & \phantom{0}2.1 & \phantom{0}5.5 \\
  & LLM-Expand & 26.8 & 68.8 & \phantom{0}4.5 \\\midrule
\multirow{2}{*}{Gemini}
  & LLM-Polish & 82.2 & \phantom{0}5.1 & \phantom{0}8.8 \\
  & LLM-Expand & 22.5 & 70.1 & \phantom{0}3.1 \\\midrule
\multirow{2}{*}{DeepSeek}
  & LLM-Polish & 84.2 & \phantom{0}6.9 & \phantom{0}7.1 \\
  & LLM-Expand & \phantom{0}6.8 & 86.0 & \phantom{0}4.8 \\
\bottomrule
\end{tabular}
\caption{\tasknameinsightprovenance~label distribution ($\%$) given different levels of LLM involvement. Rows do not sum to 100\% because unparseable judgements are excluded.}
\label{tab:source-label}
\vspace{-5mm}
\end{table}

\vspace{-3mm}
\paragraph{\tasknameinsightprovenance~Annotation.}
Reviews from \fullyhumanreview~and \fullyllmreview~can be directly labeled as \labelhumaninsight~and \labelaiinsight, respectively. In contrast, \llmparaphrase~and \llmenrichbullets~provide only coarse provenance signals, as LLMs may introduce additional insights during generation. We therefore perform sentence-level provenance annotation based on a simple intuition: \emph{the extent to which an AI-written sentence can be inferred from the human-provided content indicates how much of its underlying insight originates from the human versus the LLM}.

We first segment all reviews into sentences using spaCy\footnote{\url{https://spacy.io}} and discard sentences shorter than 10 tokens, which are predominantly section headers or list markers. For each AI-written sentence, GPT-4o judges whether its underlying meaning or insight can be inferred from the corresponding human source: the original human review for \llmparaphrase, or the human-derived bullets for \llmenrichbullets: ``\emph{If the sentence's meaning or insight can be \textbf{fully} inferred or entailed by the human source, label it \textsc{Human}; if it can be only \textbf{partially} inferred or entailed, label it \textsc{Hybrid}; if \textbf{none} of the underlying meaning or insight is supported by the human source, label it \textsc{AI}}''. The judge outputs both a provenance label and a brief justification (full prompts in Appendix~\ref{app:prompt-annot}). Table~\ref{tab:source-label} summarizes the resulting label distribution.

\vspace{-2mm}
\paragraph{Training/Dev/Test Split and Debiasing.} We first construct the splits then address  shortcuts.

\textbullet\enspace\textbf{Train-raw, Dev, and Test-raw.} Each of three generation conditions (\llmparaphrase, \llmenrichbullets, and \fullyllmreview) contributes  $\sim$50K sentences to training, sampled with a fixed seed and stratified by provenance label (\labelhumaninsight, \labelhybrid, and \labelaiinsight). For the \labelhumaninsight~class, we equally sample \fullyhumanreview~sentences and LLM-generated sentences labeled as \labelhumaninsight, balancing human-originated insights expressed in human- and LLM-written text.
The remaining sentences are split into $Dev$ and $Test$-$raw$, with balanced contributions from the three generation conditions. Table~\ref{tab:dataset_stats} (Appendix~\ref{sec:appendix-data}) reports the full statistics.

Preliminary experiments revealed two potential shortcuts: (1) models could perform well without effectively using the paper context, and (2) high performance on $Test$-$raw$ was partly driven by superficial linguistic patterns correlated with provenance labels. We address these two issues separately through \emph{training augmentation} and \emph{test debiasing}, respectively.

\textbullet\enspace\textbf{Train Augmentation ($Train$-$raw$ $\to$ $Train$).}
To encourage models to use the paper content, we augment $Train$-$raw$ with an \textit{Unrelated} class. For each $(\text{paper},\text{ review sentence})$ pair, we create an additional instance by replacing the original paper with a different but semantically most similar paper retrieved based on paper-level embedding similarity\footnote{\url{https://github.com/huggingface/sentence-transformers}}, while keeping the sentence unchanged. The new pair is labeled \textit{Unrelated}. Distinguishing these examples from the original pairs therefore requires reasoning about the relationship between the paper and the review sentence. This class is used only for training and does not appear in the evaluation sets.

\textbullet\enspace\textbf{Debiased $\mathbf{Test}$ ($Test$-$raw$ $\to$ $Debiased$-$Test$).}
To evaluate robustness to linguistic shortcuts, we construct two progressively debiased versions of $Test$-$raw$ using GPT-4o (temperature=$0.2$; prompts in Appendix~\ref{app:prompt-debias}). i) \textbf{Debiased-Test-v1 (style neutralization)} removes common surface cues, including first-person expressions, discourse markers, modal hedges, emphatic adjectives, passive voice, abbreviations, informal punctuation, and large variations in sentence length. ii) \textbf{Debiased-Test-v2 (content canonicalization)} further reduces higher-level rhetorical cues by rewriting each sentence as a short, plain statement of its core proposition, removing recommendation language, evaluative stance, numbered references, and comparison framing. Note that debiasing is applied only to \llmparaphrase, \llmenrichbullets, and \fullyllmreview; \fullyhumanreview~is left unchanged to preserve genuinely human-written text.

Table~\ref{tab:debias-audit} (Appendix~\ref{sec:appendix-data}) quantifies the progressive removal of these cues. For example, Test-v1 substantially reduces stylistic markers such as hedges, transitions, and first-person expressions, while Test-v2 further reduces structural cues such as numbered references and suggestion language. \emph{Because neither transformation guarantees a completely shortcut-free representation, we report performance on the Test-raw, Debiased-Test-v1, and Debiased-Test-v2 test sets to evaluate model robustness under progressively stronger shortcut reduction}.

\begin{figure*}[t]
\centering
\includegraphics[width=0.9\textwidth]{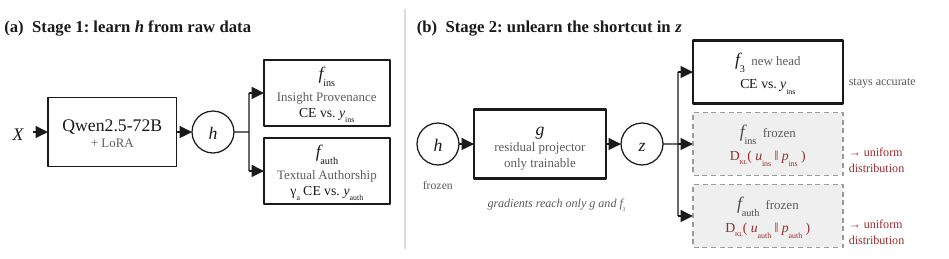}
\vspace{-5mm}
\caption{\modelname's training.}
\label{fig:architecture}
\vspace{-5mm}
\end{figure*}

\vspace{-2mm}
\section{Method: \modelname}
\vspace{-2mm}

We first distinguish two related prediction tasks:

\textbullet\enspace\tasknameinsightprovenance: $y_{\text{ins}}$ $\in$ \{\labelhumaninsight, \labelhybrid, \labelaiinsight\} --- where the insight originated. Both \fullyhumanreview~reviews and those review sentences from \llmparaphrase~and \llmenrichbullets~annotated as fully derived from human reviews are labeled \labelhumaninsight.

\textbullet\enspace\tasknameauthorship: $y_{\text{auth}}$ $\in$ \{\labelhumanwritten, \labelaiwritten\} --- who
      produced the surface form. Only reviews from \fullyhumanreview~are labeled as \labelhumanwritten.

Our goal is to learn \tasknameinsightprovenance~without relying on \tasknameauthorship~shortcuts. This is challenging because provenance and authorship are strongly correlated in the training data, causing models to exploit human-vs.-AI writing patterns rather than identify where the underlying insight originated.

We propose \modelname~(Figure~\ref{fig:architecture}), a two-stage framework. \textbf{Stage~1} jointly learns provenance and authorship, capturing genuine provenance signals alongside potential shortcuts. \textbf{Stage~2} learns a shortcut-suppressed representation by confusing the frozen Stage~1 provenance and authorship detectors while preserving provenance prediction, improving robustness across $Test$-$raw$, $Test$-$v1$, and $Test$-$v2$.


\vspace{-2mm}
\paragraph{Stage 1: Learning the Raw Representation.}

Stage~1 learns a strong representation directly from the raw training data. Two classifiers share the same representation $h$: a provenance head $f_{\text{ins}}$ predicts \tasknameinsightprovenance, while an auxiliary authorship head $f_{\text{auth}}$ predicts \tasknameauthorship. They are jointly trained with
\begin{equation}
\begin{aligned}
\mathcal{L}_{\text{S1}} ={}& \mathrm{CE}\bigl(f_{\text{ins}}(h),y_{\text{ins}}\bigr)\\
&+ \gamma_{\text{a}}\,\mathrm{CE}\bigl(f_{\text{auth}}(h),y_{\text{auth}}\bigr),
\end{aligned}
\label{eq:stage1}
\end{equation}
where $\gamma_{\text{a}}=0.5$. The authorship head serves as an auxiliary probe of authorship information encoded in $h$. For \textit{Unrelated} instances in the augmented training set, the authorship loss is masked.

At first glance, Stage~1 appears highly effective: it achieves approximately 87\% F1/accuracy for \tasknameinsightprovenance~on $Test$-$raw$, while the auxiliary head reaches approximately 96\% authorship accuracy. However, this strong authorship predictability reveals that $h$ retains substantial information about \emph{who wrote the sentence}, which is strongly correlated with provenance in the raw data but does not necessarily indicate \emph{who contributed the insight}. Thus, the strong $Test$-$raw$ provenance performance can be misleading: $h$ may encode genuine provenance information together with easily exploitable authorship and linguistic shortcuts. This motivates Stage~2, which aims to suppress these shortcut signals while preserving information needed for \tasknameinsightprovenance.

\paragraph{Stage 2: Adversarial Shortcut Unlearning.}
Our intuition is that if Stage~1 succeeds partly by exploiting shortcuts, then \emph{learning a representation that confuses its frozen detectors (i.e., pushes them toward random guessing) can suppress these shortcuts}, while a new provenance head preserves information genuinely useful for provenance prediction. Therefore, we freeze the encoder, LoRA adapter, and both
Stage~1 heads, and introduce a trainable residual projector $g$:
\vspace{-2mm}
\begin{equation}
z=g(h)=W_2\,\mathrm{LN}\bigl(\sigma(W_1h)\bigr)+h,
\vspace{-2mm}
\end{equation}
where $\sigma$ is GELU and $\mathrm{LN}$ is layer normalization.

A new \tasknameinsightprovenance~classifier $f_3$ is trained on $z$ to
preserve the target task, $\hat{y}_3=f_3(z)$, with loss
$\mathcal{L}_{\text{new-ins}}=\mathrm{CE}\bigl(f_3(z),y_{\text{ins}}\bigr)$. At the same time, we explicitly require the two frozen Stage~1 heads to
approach random guessing on $z$. Let
$p_{\text{ins}}(z)=\operatorname{softmax}(f_{\text{ins}}(z))$ and
$p_{\text{auth}}(z)=\operatorname{softmax}(f_{\text{auth}}(z))$
denote their predictive distributions. We then minimize the divergence between each frozen detector's prediction
and its corresponding uniform distribution ($u_{\text{ins}}$ and $u_{\text{auth}}$):
\vspace{-2mm}
\begin{equation}
\begin{aligned}
\mathcal{L}_{\text{confusing}}
={}& D_{\mathrm{KL}}\left(u_{\text{ins}} \,\|\, p_{\text{ins}}(z)\right)\\
&+ D_{\mathrm{KL}}\left(u_{\text{auth}} \,\|\, p_{\text{auth}}(z)\right).
\end{aligned}
\label{eq:confusion}
\vspace{-2mm}
\end{equation}
Because the Stage~1 heads are frozen, gradients from
$\mathcal{L}_{\text{confusing}}$ update only $g$, encouraging $z$ to suppress
signals that allow the original detectors to make confident predictions.
The Stage~2 objective is therefore
$
\mathcal{L}_{\text{S2}}
=
\mathcal{L}_{\text{new-ins}}
+
\lambda \mathcal{L}_{\text{confusing}}.
\label{eq:finalloss}
$ Thus, $\mathcal{L}_{\text{new-ins}}$ encourages $z$ to retain information needed for
\tasknameinsightprovenance, while $\mathcal{L}_{\text{confusing}}$ explicitly
pushes the frozen Stage~1 detectors toward random guessing.

\begin{table*}[t]
\centering
\small
\setlength{\tabcolsep}{3pt}
\begin{tabular}{l ccc | ccc | ccc}
\toprule
\multirow{2}{*}{\textbf{Model}}
  & \multicolumn{3}{c|}{\textbf{GPT}} & \multicolumn{3}{c|}{\textbf{Gemini}} & \multicolumn{3}{c}{\textbf{DeepSeek}} \\
\cmidrule(l){2-10}
  & Raw & V1 & V2 & Raw & V1 & V2 & Raw & V1 & V2\\
\midrule
  In-context GPT-4o-mini         & 48.29 & 43.83 & 43.87 & 54.48 & 47.27 & 44.22 & 41.13 & 39.73 & 41.79\\
  In-context Gemini-flash-latest & 65.66 & 57.77 & 54.00 & 63.45 & 50.89 & 47.57 & 52.58 & 47.96 & 45.76\\
  \midrule
Binoculars~\citep{DBLP:conf/icml/HansSCKSGGG24}         & 48.85 & 44.45 & 43.09 & 52.39 & 53.11 & 52.96 & 36.61 & 35.61 & 35.04 \\
Fast-DetectGPT~\citep{DBLP:conf/iclr/BaoZTY024}   & 45.37 & 44.32 & 44.78 & 49.18 & 50.71 & 51.56 & 34.96 & 34.46 & 34.88 \\
TDT~\citep{DBLP:journals/corr/abs-2508-01754}                         & 42.49 & 40.55 & 40.27 & 51.29 & 50.97 & 50.92 & 34.26 & 33.61 & 33.49 \\
MAGE~\citep{DBLP:conf/acl/LiLCBWWY0024}  & 34.95 & 40.78 & 42.32  & 52.70 & 55.91 & 51.62 & 36.40 & 42.10 & 43.85\\
RADAR~\citep{DBLP:conf/nips/HuCH23} & 40.15 & 40.20 & 40.24 & 38.90 & 39.05 & 39.30 & 41.60 & 41.75 & 42.00\\
Fakespot-RoBERTa~\citep{fakespot-roberta-2025} & 31.77 & 27.94 & 29.61 & 29.85 & 26.40 & 28.15 & 33.20 & 29.10 & 30.95\\
Hello-SimpleAI~\citep{DBLP:journals/corr/abs-2309-02731}  & 45.33 & 43.51 & 41.93 & 43.60 & 41.80 & 40.25 & 46.90 & 44.95 & 43.10\\
DeBERTa-Large-v1~\citep{DBLP:conf/iclr/HeGC23} & 47.97 & 51.02 & 49.87 & 46.15 & 49.30 & 48.05 & 49.60 & 52.75 & 51.40\\
OpenAI-RoBERTa~\citep{DBLP:journals/corr/abs-1908-09203}                        & 32.57 & 34.58 & 35.24 & 41.43 & 41.60 & 42.33 & 28.69 & 29.78 & 31.32\\
LLM-PIRATE~\citep{DBLP:conf/ndss/GohilDNSR25}                               & 43.58 & 41.48 & 42.22 & 50.97 & 48.99 & 49.39 & 35.17 & 33.99 & 34.50\\
SeqXGPT~\citep{DBLP:conf/emnlp/WangLRJZQ23}                         & 72.85 & 56.58 & 49.01 & 58.25 & 53.12 & 52.43 & 58.17 & 48.57 & 42.00 \\
DeTeCtive~\citep{DBLP:conf/nips/GuoHZZFHM24}                       & 75.45 & 68.62 & 63.08 & 66.12 & 59.74 & 58.92 & 65.63 & 59.26 & 54.63 \\
RACE~\citep{DBLP:conf/acl/LiSWYWC26} & 85.74 & 75.86 & 65.92 & 90.38 & 75.14 & \textbf{66.90} & 88.65 & 70.64 & 54.55\\
\midrule
\modelname~(ours) & \textbf{93.61}  & 79.21 & 67.16 & \textbf{92.80} & \textbf{78.05} & 66.30 & \textbf{92.21} & 70.18 & \textbf{68.45}\\
\quad w/o stage 2             & 91.00 & 77.32 & 58.31 & 74.46 & 61.74 & 56.81 & 82.27 & 64.87 & 46.90\\
\quad w/o $f_{\text{ins}}$ \& stage 2 & 89.86 & 72.19 & 55.52 & 72.13 & 60.80 & 58.11 & 76.22 & 59.94 & 44.81\\
\quad w/o $f_{\text{auth}}$ \& stage 2 & 89.62 & 74.06 & 57.76 & 71.68 & 60.22 & 57.78 & 77.76 & 62.55 & 46.06\\

\quad w/o paper $P$ in input &92.79 & \textbf{81.42} & \textbf{71.19} & 84.77 & 63.79 & 62.59 & 83.43 & 71.75 & 57.72\\
\quad w/o review contenxt  ($C_l$, $C_r$) in input & 91.29 & 78.04 & 58.12 & 74.06 & 60.56 & 55.33 & 82.00 & 65.16 & 46.25\\

\quad Replace Qwen with RoBERTa                 & 89.52 & 80.43 & 67.75 & 77.38 & 67.42 & 65.13 & 83.90  & \textbf{73.25} & 59.61 \\
\quad Replace Qwen with Longformer       & 89.49 & 80.25 & 68.38 & 77.37 & 66.85 & 65.11 & 83.59 & 72.99 & 60.04 \\
\bottomrule
\end{tabular}
\vspace{-2mm}
\caption{\modelname~vs.\ baselines. ``Raw'': Test-raw; ``V1'': Debiased-Test-v1; ``V2'': Debiased-Test-v2.}
\label{tab:effectiveness}
\vspace{-5mm}
\end{table*}

\section{Experiments}
\vspace{-2mm}

\paragraph{\modelname~Implementation.} For each target review sentence, \modelname~takes as input the tuple
$\mathcal{X}=(P, C_{\mathrm{l}}, S, C_{\mathrm{r}})$,
where $P$ is the paper, $S$ is the target sentence, and $C_{\mathrm{l}}$ and $C_{\mathrm{r}}$ denote its preceding and subsequent review sentence, respectively. In our analysis ($\mathcal{Q}_1$), we further examine the contribution of these information sources by removing the paper and/or surrounding review context.
We use Qwen2.5-72B-Instruct \citep{DBLP:journals/corr/abs-2412-15115} with LoRA
\citep{DBLP:journals/corr/abs-2106-09685} ($r{=}8$, $\alpha{=}16$, dropout $0.05$) applied to the
query and value projections. The representation $h\in\mathbb{R}^{8192}$ is
the final-layer hidden state of the last non-padding token. The target review
sentence is always retained in full, with the remaining token budget allocated
first to the paper and then to the surrounding review context.

Stage~1 trains the LoRA adapter and two prediction heads for three epochs using
AdamW ($\mathrm{lr}=2\times10^{-4}$, weight decay $0.01$). Stage~2 freezes all
Stage~1 components and trains only $g$ and $f_3$ for three epochs with
$\mathrm{lr}=10^{-4}$. The shortcut-unlearning weight $\lambda$ in
Equation~\ref{eq:finalloss} is gradually increased as
$
\lambda(p)=\frac{2}{1+e^{-\gamma p}}-1,
$
where $\gamma=10$, and $p$ denotes training progress, allowing $f_3$ to first learn the target
task before stronger shortcut-unlearning pressure is applied. 

\vspace{-2mm}
\paragraph{Baselines.}
Since no prior system directly targets \tasknameinsightprovenance, we compare
against representative \tasknameauthorship~detectors,
including MAGE \citep{DBLP:conf/acl/LiLCBWWY0024}, RADAR
\citep{DBLP:conf/nips/HuCH23}, Fakespot-RoBERTa
\citep{fakespot-roberta-2025}, Hello-SimpleAI \citep{DBLP:journals/corr/abs-2309-02731},
two DeBERTa-v3-large detectors \citep{DBLP:conf/iclr/HeGC23},
OpenAI-RoBERTa \citep{DBLP:journals/corr/abs-1908-09203}, LLM-PIRATE
\citep{DBLP:conf/ndss/GohilDNSR25}, Binoculars
\citep{DBLP:conf/icml/HansSCKSGGG24}, Fast-DetectGPT
\citep{DBLP:conf/iclr/BaoZTY024}, SeqXGPT
\citep{DBLP:conf/emnlp/WangLRJZQ23}, DeTeCtive
\citep{DBLP:conf/nips/GuoHZZFHM24}, and RACE~\citep{DBLP:conf/acl/LiSWYWC26}.
For systems that output only a binary human/AI prediction or a single
AI-written probability $P(\mathrm{AI})$, we adapt their outputs to our
three-way setting using two thresholds, $\tau_{\mathrm{H}}$ and
$\tau_{\mathrm{A}}$, tuned on $Dev$: \labelhumaninsight~($P(\mathrm{AI}) < \tau_{\mathrm{H}}$), \labelhybrid~($\tau_{\mathrm{H}} \leq P(\mathrm{AI}) < \tau_{\mathrm{A}}$), and \labelaiinsight~($P(\mathrm{AI}) \geq \tau_{\mathrm{A}}$).

\vspace{-4mm}
\paragraph{Evaluation Metrics.}
We report weighted F1 over the three provenance classes as our primary evaluation metric. All systems are evaluated on $Test$-$raw$, $Debiased$-$Test$-$v1$, and $Debiased$-$Test$-$v2$. Beyond strong performance on any individual test set, we emphasize consistently strong performance across all three, as this provides a more comprehensive measure of a model's overall competitiveness and robustness to progressively reduced textual-authorship shortcuts.

\begin{table*}[t]
\centering
\small
\setlength{\tabcolsep}{3pt}
\begin{tabular}{l ccc | ccc | ccc | c}
\toprule
\multirow{2}{*}{\textbf{Model}}
  & \multicolumn{3}{c|}{\textbf{Test on GPT}} & \multicolumn{3}{c|}{\textbf{Test on Gemini}} & \multicolumn{3}{c|}{\textbf{Test on DS}} & \multirow{2}{*}{\textbf{Mean} $\pm$ \textbf{Std.}}\\
\cmidrule(l){2-10}
   & Raw & V1 & V2 &  Raw & V1 & V2 & Raw & V1 & V2 & \\
\midrule
Train on GPT      & 93.61 & 79.21 & 67.16 & 90.18 & 75.67 & 62.33 & 91.21 & 78.69 & 62.12 & 77.80 $\pm$ 13.21\\
Train on Gemini   & 89.83 & 75.83 & 56.24 & 92.80 & 78.05 & 66.30 & 78.25 & 69.54 & 53.24 & 73.34 $\pm$ 16.22\\
Train on DeepSeek & 81.55 & 74.71 & 56.25 & 79.85 & 70.36 & 52.13 & 92.21 & 70.18 & 68.45 & 71.74 $\pm$ 14.71\\
Train on mix      & 93.54 & 80.01 & 66.23 & 84.23 & 72.36 & 60.34 & 86.15 & 76.89 & 59.97 & 75.52 $\pm$ 13.44\\
\bottomrule
\end{tabular}
\vspace{-2mm}
\caption{Cross-LLM transfer and robustness (weighted F1).}
\label{tab:robustness}
\vspace{-5mm}
\end{table*}

\subsection{Main Results}

\paragraph{\modelname~vs.\ baselines.}
Table~\ref{tab:effectiveness} demonstrates that \tasknameinsightprovenance~poses a fundamentally different challenge from conventional \tasknameauthorship~detection. First, existing authorship detectors transfer poorly to our task. Even the strongest baselines exhibit substantial degradation as textual-authorship shortcuts are progressively removed from $Test$-$raw$ to $Debiased$-$Test$-$v1$ and $Debiased$-$Test$-$v2$. For example, RACE drops from $85.74$ to $65.92$ on GPT, from $90.38$ to $66.90$ on Gemini, and from $88.65$ to $54.55$ on DeepSeek. This suggests that signals effective for identifying \emph{who wrote the text} are insufficient for determining \emph{who contributed the underlying insight}.

Second, directly prompting strong LLMs with in-context examples does not solve the task either. Both GPT-4o-mini and Gemini-Flash achieve substantially lower performance than the strongest trained systems, particularly on the debiased test sets. This indicates that insight provenance is not readily recoverable from general-purpose LLM reasoning alone and motivates learning representations specifically for separating intellectual provenance from surface-generation cues.

Third, \modelname~achieves consistently strong performance across generators and test conditions, outperforming the competing baselines in most settings. The advantage is particularly pronounced on $Debiased$-$Test$-$v2$, where textual-authorship cues are most aggressively controlled: \modelname~achieves $67.16$, $66.30$, and $68.45$ weighted F1 on GPT, Gemini, and DeepSeek, respectively, substantially improving over most existing textual-authorship detectors while remaining competitive with or exceeding the strongest baseline.

Our \textbf{ablations} further validate the proposed two-stage design. Removing Stage~2 causes substantial degradation, especially on the most strongly debiased $V2$ sets, showing that explicitly suppressing textual-authorship shortcuts is critical for robust insight-provenance detection. Finally, the effectiveness of \modelname~is not tied to a particular backbone: replacing Qwen2.5-72B-Instruct with the classifier-oriented RoBERTa \citep{liu2019robertarobustlyoptimizedbert} or Longformer \citep{beltagy2020longformerlongdocumenttransformer} retains strong performance and generally remains substantially stronger than conventional textual-authorship detectors, particularly under debiasing. Overall, these results suggest that the gains arise primarily from the proposed provenance-oriented training framework rather than from the choice of backbone model.

\vspace{-3mm}
\paragraph{Robustness  across different LLMs' test sets}
A desirable insight-provenance detector should not be narrowly tied to reviews generated by a particular LLM, since the underlying task concerns \emph{who contributed the insight}, rather than which model produced the text. Table~\ref{tab:robustness} therefore evaluates \modelname~under cross-LLM transfer among GPT, Gemini, and DeepSeek. When trained on a single LLM, performance generally varies across test generators, with larger degradation under the more aggressively debiased settings, indicating that generator-specific signals can still be learned despite our provenance-oriented training. Importantly, training on the mixture of multiple LLMs yields more balanced performance across test generators, suggesting that generator diversity during training can reduce reliance on model-specific characteristics and improve cross-LLM robustness.

We emphasize that this experiment does not establish universal generalization to arbitrary or future unseen LLMs: all three LLM families represented at test time also contribute to the mixed training setting. Rather, the results provide evidence for a practical direction toward more general insight-provenance detection: increasing generator diversity during training can encourage the detector to capture signals that transfer across different LLM-generated reviews, instead of specializing to the  characteristics of a single LLM.

\vspace{-2mm}
\section{Analysis}
\vspace{-2mm}

\begin{table*}[t]
\centering\small
\begin{tabular}{llcccc}
\toprule
& & \multicolumn{4}{c}{Condition (\% human-attributable)} \\
\cmidrule(lr){3-6}
Dimension & Semantic & AI (0\%) & H$_{\mathrm{low}}$ (27\%) & H$_{\mathrm{high}}$ (91\%) & H (100\%) \\
\midrule
Specificity       & concrete referent          & \phantom{0}4.3 & 14.4 & 23.1 & 23.4 \\
Verifiability     & explicit referent          & 14.3 & 18.4 & 27.3 & 30.7 \\
Fabricated ref.   & of all claims              & 0.28 & 2.58 & 4.14 & 4.93 \\
Correctness       & error rate among all verifiable  & 25.4 & 49.3 & 43.0 & 33.8 \\
Actionability     & specific action            & 21.0 & 20.0 & 15.0 & 14.0 \\
Novelty           & grading by Gemini          & 16.5 & 11.8 & 19.8 & 19.8 \\
Transplantability & holding on unrelated paper & 24.0 & 20.8 & 23.7 & 21.5 \\
\bottomrule
\end{tabular}
\vspace{-2mm}
\caption{Quality and grounding characteristics of review insights across different levels of human involvement (AI: \fullyllmreview; H$_{low}$: \llmenrichbullets; H$_{high}$: \llmparaphrase; H: \fullyhumanreview).}
\label{tab:q5-dims}
\vspace{-3mm}
\end{table*}

\paragraph{$\mathcal{Q}_1$: Does human involvement make the review better, or simply different?}
We compare four conditions, ordered by the share of human-attributable content:
\fullyllmreview~($0\%$), \llmenrichbullets~($27\%$), \llmparaphrase~($91\%$), and
\fullyhumanreview~($100\%$). Each insight is scored on seven dimensions by three independent
judges, gpt-4o-mini, gemini-3.5-flash-lite and a locally hosted Qwen2.5-72B-Instruct. Because the dimensions differ in how reliably they can be measured, we group them into
three families. Three are \emph{extractive},
detecting a surface property rather than evaluating the claim: whether the claim names a concrete referent such as a table, equation, section, or cited work (\textbf{specificity}); whether that referent is explicit enough to locate (\textbf{verifiability}); and whether a named referent fails to resolve (\textbf{fabricated reference}). Three are \emph{judged}: \textbf{correctness},
\textbf{actionability}, and \textbf{novelty} against a common reference set of critiques for the same paper. \textbf{Transplantability} is measured by intervention: we move each claim to an unrelated paper and ask whether it still holds, giving a rubric-free measure of how paper-specific it is. Correctness is reported over verifiable claims only, since a claim with no checkable referent can be neither confirmed nor refuted. 

The evidence in Table~\ref{tab:q5-dims} supports \emph{different} more strongly than \emph{better}. We do not observe a reliable monotonic relationship between the amount of human involvement and general review quality. Moreover, LLM judges show very low agreement on subjective dimensions such as correctness, verifiability, and specificity, making strong quality conclusions difficult without human evaluation.

More reliable measurements instead reveal systematic differences in how insights are grounded. Specific and verifiable references become more common as human involvement increases: concrete referents increase from $4.3\%$ in AI-origin insights to $23.4\%$ in human-origin insights, and explicit verifiable referents increase from $14.3\%$ to $30.7\%$. Thus, human involvement changes the \emph{source and grounding} of review insights more clearly than it changes their overall quality.

\begin{table}[t]
\centering\small
\setlength{\tabcolsep}{3pt}
\begin{tabular}{lccc}
\toprule
& Human & Hybrid & AI \\
\midrule
Field-generic & 18.0 & 29.8 & \textbf{44.5} \\
                          & {\scriptsize[16.4, 19.7]} & {\scriptsize[26.4, 33.9]} & {\scriptsize[41.4, 47.6]} \\
Paper-explicit            & 21.9 & 28.5 & \textbf{38.6} \\
                          & {\scriptsize[19.8, 23.9]} & {\scriptsize[24.7, 32.3]} & {\scriptsize[35.7, 41.7]} \\
Paper-inferable           & \phantom{0}3.2 & \phantom{0}2.9 & \phantom{0}2.7 \\
                          & {\scriptsize[2.6, 4.0]} & {\scriptsize[1.7, 4.2]} & {\scriptsize[1.8, 3.7]} \\
External knowledge        & \textbf{28.0} & 13.0 & \phantom{0}2.7 \\
                          & {\scriptsize[25.8, 30.2]} & {\scriptsize[10.1, 15.9]} & {\scriptsize[1.9, 3.8]} \\
Experience-based          & \textbf{28.9} & 25.3 & 11.1 \\
                 judgement         & {\scriptsize[26.9, 31.1]} & {\scriptsize[21.8, 28.8]} & {\scriptsize[9.3, 13.1]} \\
\bottomrule
\end{tabular}
\vspace{-2mm}
\caption{Provenance category distribution (\%).}
\label{tab:insightdistancetopaper}
\vspace{-6mm}
\end{table}

\begin{table*}[t]
\centering\small
\setlength{\tabcolsep}{3.5pt}
\begin{tabular}{llrccc}
\toprule
& & $n$ & $X=(P,S)$ & $X=(P, C_{\mathrm{l}}, S, C_{\mathrm{r}})$ & $\Delta$ \\
\midrule
\multicolumn{6}{l}{\textit{(a) How does left–right provenance agreement affect performance?}} \\
\multicolumn{2}{l}{All \{$S$\}}                 & 81{,}900 & 0.8889 & 0.9246 & $+0.036$ \\
\multicolumn{2}{l}{C1: \{$S$\} that have both $C_{\mathrm{l}}$ and $C_{\mathrm{r}}$ }       & 50{,}659 & 0.8817 & 0.9174 & $+0.036$ \\
\multicolumn{2}{l}{\quad $C_{\mathrm{l}}^{y_{\mathrm{ins}}}=C_{\mathrm{r}}^{y_{\mathrm{ins}}}$}   & 41{,}633 & 0.9232 & \textbf{0.9645} & $+0.041$ \\
\multicolumn{2}{l}{\quad $C_{\mathrm{l}}^{y_{\mathrm{ins}}}\neq C_{\mathrm{r}}^{y_{\mathrm{ins}}}$} & \phantom{0}9{,}026 & \textbf{0.5460} & 0.5331 & $\mathbf{-0.013}$ \\
\midrule
\multicolumn{6}{l}{\textit{(b) Does context help through discourse relations or provenance cues?}} \\
C1 & real context $C_{\mathrm{l}}$ and $C_{\mathrm{r}}$ & 50{,}659 &  & 0.9174 & [0.915, 0.920] \\
C2 & donor context, \emph{same} class & 50{,}659 &  & \textbf{0.9285} & [0.926, 0.931] \\
C3 & donor context, \emph{other} class& 50{,}659 &  & \textbf{0.3159} & [0.312, 0.320] \\
\bottomrule
\end{tabular}
\vspace{-2mm}
\caption{Surrounding review sentences primarily provide a local provenance prior.}
\label{tab:context-mechanism}
\vspace{-4mm}
\end{table*}

\paragraph{$\mathcal{Q}_2$: Why is intellectual authorship identifiable at all?}

The previous analyses suggest that \tasknameinsightprovenance~is fundamentally a question of \emph{provenance}: how far an insight travels beyond the information already available to the reviewer from the submission. We therefore measure the \emph{provenance distance} of each insight by asking how much information from the paper an independent generator needs before it can reproduce a semantically equivalent observation. We label five sources of insight using GPT-4o: \textbf{field-generic}, which requires no access to the paper; \textbf{paper-explicit}, which is already stated in the submission; \textbf{paper-inferable}, which requires reasoning over the paper's evidence; \textbf{external knowledge}, which requires information such as prior work, benchmarks, or community standards; and \textbf{Experience-based judgment},  reflecting an evaluation informed by the reviewer’s experience or expertise rather than directly supplied by the paper.

The resulting distributions in Table~\ref{tab:insightdistancetopaper} reveal a strong monotonic separation among intellectual contributors. For \labelaiinsight, $44.5\%$ are field-generic and another $38.6\%$ are paper-explicit. Thus, $83.1\%$ require nothing beyond generic review knowledge and information already supplied by the submission. The corresponding proportion for \labelhumaninsight~is only $39.9\%$.

The opposite pattern appears for information beyond the submission. External knowledge accounts for $28.0\%$ of \labelhumaninsight~but only $2.7\%$ of \labelaiinsight, while reviewer judgment accounts for $28.9\%$ and $11.1\%$, respectively. Interestingly, paper-inferable insights are rare for all three classes ($2.7$--$3.2\%$), suggesting that the main distinction is not that humans perform substantially deeper reasoning over tables or proofs. Rather, humans are substantially more likely to contribute information and judgment that were not available in the submission in the first place.

Taken together, these findings explain why \tasknameinsightprovenance~remains detectable after linguistic normalization. Rewriting can remove first-person language, hedging, formatting, verbosity, and other stylistic cues, but it cannot easily change \emph{where an insight came from}. An observation copied or generated from the submission remains grounded in the submission; an observation based on external literature or independent reviewer judgment retains that provenance even when its wording is rewritten.


\paragraph{$\mathcal{Q}_3$: Regarding the input choice $\mathcal{X}=(P, C_{\mathrm{left}}, S, C_{\mathrm{right}})$, what information do the paper $P$ and surrounding review sentences (i.e., $C_{\mathrm{left}}$ and $C_{\mathrm{right}}$) provide?}\enspace\\

\vspace{-5mm}
\textbf{What does the paper add?}
We found two complementary signals. First, it serves as a
\emph{provenance reference}, as established in $\mathcal{Q}_2$: comparing an insight against the submission
reveals whether it can be explained by information already available in the
paper or requires information beyond it. This distinction is strongly
associated with provenance: external knowledge and experiential judgment together
account for $56.9\%$ of \labelhumaninsight~but only $13.8\%$ of
\labelaiinsight, whereas field-generic and paper-explicit information accounts
for $83.1\%$ of \labelaiinsight~but only $39.9\%$ of
\labelhumaninsight. Second, the paper provides \emph{semantic context} for
interpreting what a critique refers to. Human insights are substantially more
likely to contain concrete ($23.4\%$ vs.\ $4.3\%$) and explicitly verifiable
($30.7\%$ vs.\ $14.3\%$) referents than AI insights; grounding such references
in the paper helps identify the specific claims, experiments, or results being
evaluated rather than relying on the review sentence alone.



\textbf{What do surrounding review sentences add?}
Adding neighboring review sentences produces a larger improvement, but our analysis suggests that much of this gain comes from \emph{local provenance consistency}, rather than from understanding the discourse relation between adjacent sentences.

Sentence provenance is strongly autocorrelated within a review. As an oracle diagnostic, using $x^{y_{\mathrm{ins}}}$ to denote the gold label of $x$, the diagonal probabilities of $P(S^{y_{\mathrm{ins}}} \mid C_{\mathrm{left}}^{y_{\mathrm{ins}}})$ are $0.900$, $0.685$, and $0.960$ for \labelhumaninsight, \labelhybrid, and \labelaiinsight~sentences, respectively. The label entropy decreases from $1.418$ bits marginally to $0.529$ bits given the left neighbor and $0.383$ bits given both neighbors. These statistics use gold labels only to characterize the dataset; gold neighboring labels are \emph{not} available to the classifier at inference time.

The context intervention in Table~\ref{tab:context-mechanism} provides stronger evidence. On examples whose two neighbors have the same provenance, full context improves  from $0.9232$ to $0.9645$. When neighboring provenance disagrees, however, performance decreases from $0.5460$ to $0.5331$ (block (a) of Table~\ref{tab:context-mechanism}). More strikingly, replacing the true neighboring sentences with sentences drawn from an unrelated review but having the same label produces comparable performance ($0.9285$ vs. $0.9174$ with the real context), whereas replacing them with text from another  class reduces performance to $0.3159$ (block (b) of the table).

These results suggest that neighboring sentences primarily provide a \emph{local provenance prior}: text surrounding a target sentence often carries signals about the contributor responsible for nearby insights, and contributor identity tends to persist across neighboring sentences. Importantly, this does not imply that the model observes neighboring gold labels. Rather, the intervention suggests that provenance-related properties of the neighboring \emph{text} can influence prediction of the target sentence. We therefore view surrounding context as complementary to paper grounding: the paper helps determine \emph{where the target insight could have come from}, while neighboring sentences provide evidence about \emph{who is contributing insights in the local region of the review}.

\section{Conclusion}

This work shifts AI attribution from \emph{who wrote the words} to \emph{who contributed the ideas}. Our results show that insight provenance remains detectable beyond surface authorship cues, partly through differences in how human and AI insights are grounded. As human--AI collaboration deepens, insight provenance provides a complementary basis for tracing contribution and accountability.

\bibliography{tacl2021}

@article{DBLP:journals/corr/abs-2410-21276,
  author       = {OpenAI},
  title        = {GPT-4o System Card},
  journal      = {CoRR},
  volume       = {abs/2410.21276},
  year         = {2024},
  url          = {https://doi.org/10.48550/arXiv.2410.21276},
  doi          = {10.48550/ARXIV.2410.21276},
  eprinttype   = {arXiv},
  eprint       = {2410.21276},
  bibsource    = {dblp computer science bibliography, https://dblp.org}
}

@article{DBLP:journals/corr/abs-2412-19437,
  author       = {DeepSeek{-}AI},
  title        = {DeepSeek-V3 Technical Report},
  journal      = {CoRR},
  volume       = {abs/2412.19437},
  year         = {2024},
  url          = {https://doi.org/10.48550/arXiv.2412.19437},
  doi          = {10.48550/ARXIV.2412.19437},
  eprinttype   = {arXiv},
  eprint       = {2412.19437},
  bibsource    = {dblp computer science bibliography, https://dblp.org}
}

@inproceedings{DBLPAlchokrSSLK24,
  author       = {Rand Alchokr and
                  Evelyn Starzew and
                  Gunter Saake and
                  Thomas Leich and
                  Jacob Kr{\"{u}}ger},
  editor       = {Samuel Kai{-}Wah Chu and
                  Xiao Hu},
  title        = {The Impact of {AI} Language Models on Scientific Writing and Scientific
                  Peer Reviews: {A} Systematic Literature Review},
  booktitle    = {Proceedings of the 24th {ACM/IEEE} Joint Conference on Digital Libraries,
                  {JCDL} 2024, Hong Kong, China, December 16-20, 2024},
  pages        = {32:1--32:6},
  publisher    = {{ACM}},
  year         = {2024},
  url          = {https://doi.org/10.1145/3677389.3702508},
  doi          = {10.1145/3677389.3702508},
  bibsource    = {dblp computer science bibliography, https://dblp.org}
}

@article{DBLP01268,
  author       = {Weixin Liang and
                  Yaohui Zhang and
                  Zhengxuan Wu and
                  Haley Lepp and
                  Wenlong Ji and
                  Xuandong Zhao and
                  Hancheng Cao and
                  Sheng Liu and
                  Siyu He and
                  Zhi Huang and
                  Diyi Yang and
                  Christopher Potts and
                  Christopher D. Manning and
                  James Y. Zou},
  title        = {Mapping the Increasing Use of LLMs in Scientific Papers},
  journal      = {CoRR},
  volume       = {abs/2404.01268},
  year         = {2024},
  url          = {https://doi.org/10.48550/arXiv.2404.01268},
  doi          = {10.48550/ARXIV.2404.01268},
  eprinttype   = {arXiv},
  eprint       = {2404.01268},
  bibsource    = {dblp computer science bibliography, https://dblp.org}
}

@inproceedings{
saha2026policies,
title={Policies Permitting {LLM} Use for Polishing Peer Reviews Are Currently Not Enforceable},
author={Rounak Saha and Gurusha Juneja and Dayita Chaudhuri and Naveeja Sajeevan and Nihar B Shah and Danish Pruthi},
booktitle={Forty-third International Conference on Machine Learning},
year={2026},
url={https://openreview.net/forum?id=znFIR7WFGt}
}

@inproceedings{
yu2026is,
title={Is Your Paper Being Reviewed by an {LLM}? Benchmarking {AI} Text Detection in Peer Review},
author={Sungduk Yu and Man Luo and Avinash Madasu and Vasudev Lal and Phillip Howard},
booktitle={The Fourteenth International Conference on Learning Representations},
year={2026},
url={https://openreview.net/forum?id=HyZwf1rt4s}
}

@inproceedings{kumar-etal-2024-quis,
    title = "`Quis custodiet ipsos custodes?' Who will watch the watchmen? On Detecting {AI}-generated peer-reviews",
    author = "Kumar, Sandeep  and
      Sahu, Mohit  and
      Gacche, Vardhan  and
      Ghosal, Tirthankar  and
      Ekbal, Asif",
    editor = "Al-Onaizan, Yaser  and
      Bansal, Mohit  and
      Chen, Yun-Nung",
    booktitle = "Proceedings of the 2024 Conference on Empirical Methods in Natural Language Processing",
    month = nov,
    year = "2024",
    address = "Miami, Florida, USA",
    publisher = "Association for Computational Linguistics",
    url = "https://aclanthology.org/2024.emnlp-main.1262/",
    doi = "10.18653/v1/2024.emnlp-main.1262",
    pages = "22663--22679"
}

@article{van2026beyond,
  title={Beyond AI disclosure: Claim accountability and responsible research in scholarly publishing},
  author={Van Zoonen, Ward and Tursunbayeva, Aizhan and Morgan-Thomas, Anna},
  journal={European Management Journal},
  year={2026},
  publisher={Elsevier}
}

@article{leimeister2026chatgpt,
  title={From ‘ChatGPT et al.’to generative artificial intelligence (GenAI) in peer review: Safeguarding trust while modernising scholarly publishing},
  author={Leimeister, Jan Marco and Aaltonen, Aleksi and Currie, Wendy and Schlagwein, Daniel},
  journal={Journal of Information Technology},
  volume={41},
  number={2},
  pages={76--80},
  year={2026},
  publisher={SAGE Publications Sage UK: London, England}
}

@article{DBLP:journals/corr/abs-2312-11805,
  author       = {Gemini Team},
  title        = {Gemini: {A} Family of Highly Capable Multimodal Models},
  journal      = {CoRR},
  volume       = {abs/2312.11805},
  year         = {2023},
  url          = {https://doi.org/10.48550/arXiv.2312.11805},
  doi          = {10.48550/ARXIV.2312.11805},
  eprinttype   = {arXiv},
  eprint       = {2312.11805},
  bibsource    = {dblp computer science bibliography, https://dblp.org}
}

@inproceedings{DBLP:conf/emnlp/WangLRJZQ23,
  author       = {Pengyu Wang and
                  Linyang Li and
                  Ke Ren and
                  Botian Jiang and
                  Dong Zhang and
                  Xipeng Qiu},
  editor       = {Houda Bouamor and
                  Juan Pino and
                  Kalika Bali},
  title        = {SeqXGPT: Sentence-Level AI-Generated Text Detection},
  booktitle    = {Proceedings of the 2023 Conference on Empirical Methods in Natural
                  Language Processing, {EMNLP} 2023, Singapore, December 6-10, 2023},
  pages        = {1144--1156},
  publisher    = {Association for Computational Linguistics},
  year         = {2023},
  url          = {https://doi.org/10.18653/v1/2023.emnlp-main.73},
  doi          = {10.18653/V1/2023.EMNLP-MAIN.73},
  bibsource    = {dblp computer science bibliography, https://dblp.org}
}

@inproceedings{DBLP:conf/naacl/ZhangGCHHSZLFWS24,
  author       = {Qihui Zhang and
                  Chujie Gao and
                  Dongping Chen and
                  Yue Huang and
                  Yixin Huang and
                  Zhenyang Sun and
                  Shilin Zhang and
                  Weiye Li and
                  Zhengyan Fu and
                  Yao Wan and
                  Lichao Sun},
  editor       = {Kevin Duh and
                  Helena G{\'{o}}mez{-}Adorno and
                  Steven Bethard},
  title        = {LLM-as-a-Coauthor: Can Mixed Human-Written and Machine-Generated Text
                  Be Detected?},
  booktitle    = {Findings of the Association for Computational Linguistics: {NAACL}
                  2024, Mexico City, Mexico, June 16-21, 2024},
  series       = {Findings of {ACL}},
  volume       = {{NAACL} 2024},
  pages        = {409--436},
  publisher    = {Association for Computational Linguistics},
  year         = {2024},
  url          = {https://doi.org/10.18653/v1/2024.findings-naacl.29},
  doi          = {10.18653/V1/2024.FINDINGS-NAACL.29},
  bibsource    = {dblp computer science bibliography, https://dblp.org}
}

@inproceedings{DBLP:conf/acl/SuWWZL25,
  author       = {Zhixiong Su and
                  Yichen Wang and
                  Herun Wan and
                  Zhaohan Zhang and
                  Minnan Luo},
  editor       = {Wanxiang Che and
                  Joyce Nabende and
                  Ekaterina Shutova and
                  Mohammad Taher Pilehvar},
  title        = {HACo-Det: {A} Study Towards Fine-Grained Machine-Generated Text Detection
                  under Human-AI Coauthoring},
  booktitle    = {Proceedings of the 63rd Annual Meeting of the Association for Computational
                  Linguistics (Volume 1: Long Papers), {ACL} 2025, Vienna, Austria,
                  July 27 - August 1, 2025},
  pages        = {22015--22036},
  publisher    = {Association for Computational Linguistics},
  year         = {2025},
  url          = {https://doi.org/10.18653/v1/2025.acl-long.1069},
  doi          = {10.18653/V1/2025.ACL-LONG.1069},
  bibsource    = {dblp computer science bibliography, https://dblp.org}
}

@inproceedings{DBLP:conf/nips/HeZCML25,
  author       = {Yongxin He and
                  Shan Zhang and
                  Yixuan Cao and
                  Lei Ma and
                  Ping Luo},
  editor       = {Danielle Belgrave and
                  Cheng Zhang and
                  Laura N. Montoya and
                  Hsuan{-}Tien Lin and
                  Razvan Pascanu and
                  Piotr Koniusz and
                  Marzyeh Ghassemi and
                  Nancy Chen and
                  Iv{\'{a}}n Vladimir Meza Ru{\'{\i}}z and
                  Arturo Loaiza{-}Bonilla},
  title        = {DETree: DEtecting Human-AI Collaborative Texts via Tree-Structured
                  Hierarchical Representation Learning},
  booktitle    = {Advances in Neural Information Processing Systems 38: Annual Conference
                  on Neural Information Processing Systems 2025, NeurIPS 2025, San Diego,
                  CA, USA, December 2-7, 2025 / Mexico City, Mexico, November 30 - December
                  5, 2025},
  year         = {2025},
  url          = {http://papers.nips.cc/paper\_files/paper/2025/hash/d6f8517fceeca1e2cd61721dff786c14-Abstract-Conference.html},
  bibsource    = {dblp computer science bibliography, https://dblp.org}
}

@inproceedings{DBLP:conf/eacl/TaVHLNNWNS26,
  author       = {Minh Ngoc Ta and
                  Dong Cao Van and
                  Duc{-}Anh Hoang and
                  Minh Le{-}Anh and
                  Truong Nguyen and
                  My Anh Tran Nguyen and
                  Yuxia Wang and
                  Preslav Nakov and
                  Dinh Viet Sang},
  editor       = {Vera Demberg and
                  Kentaro Inui and
                  Llu{\'{\i}}s Marquez},
  title        = {{FAID:} Fine-grained AI-generated Text Detection using Multi-task
                  Auxiliary and Multi-level Contrastive Learning},
  booktitle    = {Proceedings of the 19th Conference of the European Chapter of the
                  Association for Computational Linguistics, {EACL} 2026 - Volume 1:
                  Long Papers, Rabat, Morocco, March 24-29, 2026},
  pages        = {3275--3296},
  publisher    = {Association for Computational Linguistics},
  year         = {2026},
  url          = {https://doi.org/10.18653/v1/2026.eacl-long.151},
  doi          = {10.18653/V1/2026.EACL-LONG.151},
  bibsource    = {dblp computer science bibliography, https://dblp.org}
}

@inproceedings{DBLP:conf/icml/Mitchell0KMF23,
  author       = {Eric Mitchell and
                  Yoonho Lee and
                  Alexander Khazatsky and
                  Christopher D. Manning and
                  Chelsea Finn},
  editor       = {Andreas Krause and
                  Emma Brunskill and
                  Kyunghyun Cho and
                  Barbara Engelhardt and
                  Sivan Sabato and
                  Jonathan Scarlett},
  title        = {DetectGPT: Zero-Shot Machine-Generated Text Detection using Probability
                  Curvature},
  booktitle    = {International Conference on Machine Learning, {ICML} 2023, 23-29 July
                  2023, Honolulu, Hawaii, {USA}},
  series       = {Proceedings of Machine Learning Research},
  volume       = {202},
  pages        = {24950--24962},
  publisher    = {{PMLR}},
  year         = {2023},
  url          = {https://proceedings.mlr.press/v202/mitchell23a.html},
  bibsource    = {dblp computer science bibliography, https://dblp.org}
}

@inproceedings{DBLP:conf/iclr/BaoZTY024,
  author       = {Guangsheng Bao and
                  Yanbin Zhao and
                  Zhiyang Teng and
                  Linyi Yang and
                  Yue Zhang},
  title        = {Fast-DetectGPT: Efficient Zero-Shot Detection of Machine-Generated
                  Text via Conditional Probability Curvature},
  booktitle    = {The Twelfth International Conference on Learning Representations,
                  {ICLR} 2024, Vienna, Austria, May 7-11, 2024},
  publisher    = {OpenReview.net},
  year         = {2024},
  url          = {https://openreview.net/forum?id=Bpcgcr8E8Z},
  bibsource    = {dblp computer science bibliography, https://dblp.org}
}

@inproceedings{DBLP:conf/icml/HansSCKSGGG24,
  author       = {Abhimanyu Hans and
                  Avi Schwarzschild and
                  Valeriia Cherepanova and
                  Hamid Kazemi and
                  Aniruddha Saha and
                  Micah Goldblum and
                  Jonas Geiping and
                  Tom Goldstein},
  editor       = {Ruslan Salakhutdinov and
                  Zico Kolter and
                  Katherine A. Heller and
                  Adrian Weller and
                  Nuria Oliver and
                  Jonathan Scarlett and
                  Felix Berkenkamp},
  title        = {Spotting LLMs With Binoculars: Zero-Shot Detection of Machine-Generated
                  Text},
  booktitle    = {Forty-first International Conference on Machine Learning, {ICML} 2024,
                  Vienna, Austria, July 21-27, 2024},
  series       = {Proceedings of Machine Learning Research},
  volume       = {235},
  pages        = {17519--17537},
  publisher    = {{PMLR} / OpenReview.net},
  year         = {2024},
  url          = {https://proceedings.mlr.press/v235/hans24a.html},
  bibsource    = {dblp computer science bibliography, https://dblp.org}
}

@article{DBLP:journals/corr/abs-2301-07597,
  author       = {Biyang Guo and
                  Xin Zhang and
                  Ziyuan Wang and
                  Minqi Jiang and
                  Jinran Nie and
                  Yuxuan Ding and
                  Jianwei Yue and
                  Yupeng Wu},
  title        = {How Close is ChatGPT to Human Experts? Comparison Corpus, Evaluation,
                  and Detection},
  journal      = {CoRR},
  volume       = {abs/2301.07597},
  year         = {2023},
  url          = {https://doi.org/10.48550/arXiv.2301.07597},
  doi          = {10.48550/ARXIV.2301.07597},
  eprinttype   = {arXiv},
  eprint       = {2301.07597},
  bibsource    = {dblp computer science bibliography, https://dblp.org}
}

@inproceedings{DBLP:conf/acl/LiLCBWWY0024,
  author       = {Yafu Li and
                  Qintong Li and
                  Leyang Cui and
                  Wei Bi and
                  Zhilin Wang and
                  Longyue Wang and
                  Linyi Yang and
                  Shuming Shi and
                  Yue Zhang},
  editor       = {Lun{-}Wei Ku and
                  Andre Martins and
                  Vivek Srikumar},
  title        = {{MAGE:} Machine-generated Text Detection in the Wild},
  booktitle    = {Proceedings of the 62nd Annual Meeting of the Association for Computational
                  Linguistics (Volume 1: Long Papers), {ACL} 2024, Bangkok, Thailand,
                  August 11-16, 2024},
  pages        = {36--53},
  publisher    = {Association for Computational Linguistics},
  year         = {2024},
  url          = {https://doi.org/10.18653/v1/2024.acl-long.3},
  doi          = {10.18653/V1/2024.ACL-LONG.3},
  bibsource    = {dblp computer science bibliography, https://dblp.org}
}

@inproceedings{DBLP:conf/nips/HuCH23,
  author       = {Xiaomeng Hu and
                  Pin{-}Yu Chen and
                  Tsung{-}Yi Ho},
  editor       = {Alice Oh and
                  Tristan Naumann and
                  Amir Globerson and
                  Kate Saenko and
                  Moritz Hardt and
                  Sergey Levine},
  title        = {{RADAR:} Robust AI-Text Detection via Adversarial Learning},
  booktitle    = {Advances in Neural Information Processing Systems 36: Annual Conference
                  on Neural Information Processing Systems 2023, NeurIPS 2023, New Orleans,
                  LA, USA, December 10 - 16, 2023},
  year         = {2023},
  url          = {http://papers.nips.cc/paper\_files/paper/2023/hash/30e15e5941ae0cdab7ef58cc8d59a4ca-Abstract-Conference.html},
  bibsource    = {dblp computer science bibliography, https://dblp.org}
}

@inproceedings{DBLP:conf/emnlp/LiuZWPL023,
  author       = {Xiaoming Liu and
                  Zhaohan Zhang and
                  Yichen Wang and
                  Hang Pu and
                  Yu Lan and
                  Chao Shen},
  editor       = {Houda Bouamor and
                  Juan Pino and
                  Kalika Bali},
  title        = {CoCo: Coherence-Enhanced Machine-Generated Text Detection Under Low
                  Resource With Contrastive Learning},
  booktitle    = {Proceedings of the 2023 Conference on Empirical Methods in Natural
                  Language Processing, {EMNLP} 2023, Singapore, December 6-10, 2023},
  pages        = {16167--16188},
  publisher    = {Association for Computational Linguistics},
  year         = {2023},
  url          = {https://doi.org/10.18653/v1/2023.emnlp-main.1005},
  doi          = {10.18653/V1/2023.EMNLP-MAIN.1005},
  bibsource    = {dblp computer science bibliography, https://dblp.org}
}

@inproceedings{DBLP:conf/nips/GuoHZZFHM24,
  author       = {Xun Guo and
                  Yongxin He and
                  Shan Zhang and
                  Ting Zhang and
                  Wanquan Feng and
                  Haibin Huang and
                  Chongyang Ma},
  editor       = {Amir Globersons and
                  Lester Mackey and
                  Danielle Belgrave and
                  Angela Fan and
                  Ulrich Paquet and
                  Jakub M. Tomczak and
                  Cheng Zhang},
  title        = {DeTeCtive: Detecting AI-generated Text via Multi-Level Contrastive
                  Learning},
  booktitle    = {Advances in Neural Information Processing Systems 37: Annual Conference
                  on Neural Information Processing Systems 2024, NeurIPS 2024, Vancouver,
                  BC, Canada, December 10 - 15, 2024},
  year         = {2024},
  url          = {http://papers.nips.cc/paper\_files/paper/2024/hash/a117a3cd54b7affad04618c77c2fb18b-Abstract-Conference.html},
  bibsource    = {dblp computer science bibliography, https://dblp.org}
}

@misc{fakespot-roberta-2025,
  title        = {RoBERTa-base AI Text Detector (v1)},
  author       = {{Fakespot AI Labs}},
  year         = {2025},
  howpublished = {Hugging Face Model Hub},
  url          = {https://huggingface.co/fakespot-ai/roberta-base-ai-text-detection-v1},
  note         = {Fine-tuned RoBERTa-base model for detecting AI-generated English texts}
}

@article{DBLP:journals/corr/abs-2309-02731,
  author       = {Zhenpeng Su and
                  Xing Wu and
                  Wei Zhou and
                  Guangyuan Ma and
                  Songlin Hu},
  title        = {{HC3} Plus: {A} Semantic-Invariant Human ChatGPT Comparison Corpus},
  journal      = {CoRR},
  volume       = {abs/2309.02731},
  year         = {2023},
  url          = {https://doi.org/10.48550/arXiv.2309.02731},
  doi          = {10.48550/ARXIV.2309.02731},
  eprinttype   = {arXiv},
  eprint       = {2309.02731},
  bibsource    = {dblp computer science bibliography, https://dblp.org}
}

@inproceedings{DBLP:conf/iclr/HeGC23,
  author       = {Pengcheng He and
                  Jianfeng Gao and
                  Weizhu Chen},
  title        = {DeBERTaV3: Improving DeBERTa using ELECTRA-Style Pre-Training with
                  Gradient-Disentangled Embedding Sharing},
  booktitle    = {The Eleventh International Conference on Learning Representations,
                  {ICLR} 2023, Kigali, Rwanda, May 1-5, 2023},
  publisher    = {OpenReview.net},
  year         = {2023},
  url          = {https://openreview.net/forum?id=sE7-XhLxHA},
  bibsource    = {dblp computer science bibliography, https://dblp.org}
}

@inproceedings{DBLP:conf/ndss/GohilDNSR25,
  author       = {Vasudev Gohil and
                  Matthew DeLorenzo and
                  Veera Vishwa Achuta Sai Venkat Nallam and
                  Joey See and
                  Jeyavijayan Rajendran},
  title        = {LLMPirate: LLMs for Black-box Hardware {IP} Piracy},
  booktitle    = {32nd Annual Network and Distributed System Security Symposium, {NDSS}
                  2025, San Diego, California, USA, February 24-28, 2025},
  publisher    = {The Internet Society},
  year         = {2025},
  url          = {https://www.ndss-symposium.org/ndss-paper/llmpirate-llms-for-black-box-hardware-ip-piracy/},
  bibsource    = {dblp computer science bibliography, https://dblp.org}
}

@article{DBLP:journals/corr/abs-2508-01754,
  author       = {Alva West and
                  Yixuan Weng and
                  Minjun Zhu and
                  Luodan Zhang and
                  Zhen Lin and
                  Guangsheng Bao and
                  Yue Zhang},
  title        = {AI-Generated Text is Non-Stationary: Detection via Temporal Tomography},
  journal      = {CoRR},
  volume       = {abs/2508.01754},
  year         = {2025},
  url          = {https://doi.org/10.48550/arXiv.2508.01754},
  doi          = {10.48550/ARXIV.2508.01754},
  eprinttype   = {arXiv},
  eprint       = {2508.01754},
  bibsource    = {dblp computer science bibliography, https://dblp.org}
}

@inproceedings{DBLP:conf/acl/LiSWYWC26,
  author       = {Yang Li and
                  Qiang Sheng and
                  Zhengjia Wang and
                  Yehan Yang and
                  Danding Wang and
                  Juan Cao},
  editor       = {Maria Liakata and
                  Viviane P. Moreira and
                  Jiajun Zhang and
                  David Jurgens},
  title        = {Beyond the Final Actor: Modeling the Dual Roles of Creator and Editor
                  for Fine-Grained LLM-Generated Text Detection},
  booktitle    = {Proceedings of the 64th Annual Meeting of the Association for Computational
                  Linguistics (Volume 1: Long Papers), {ACL} 2026, San Diego, California,
                  United States, July 2-7, 2026},
  pages        = {5188--5203},
  publisher    = {Association for Computational Linguistics},
  year         = {2026},
  url          = {https://doi.org/10.18653/v1/2026.acl-long.235},
  doi          = {10.18653/V1/2026.ACL-LONG.235},
  bibsource    = {dblp computer science bibliography, https://dblp.org}
}

@article{DBLP:journals/corr/abs-2412-15115,
  author       = {An Yang and
                  Baosong Yang and
                  Beichen Zhang and
                  Binyuan Hui and
                  Bo Zheng and
                  Bowen Yu and
                  Chengyuan Li and
                  Dayiheng Liu and
                  Fei Huang and
                  Haoran Wei and
                  Huan Lin and
                  Jian Yang and
                  Jianhong Tu and
                  Jianwei Zhang and
                  Jianxin Yang and
                  Jiaxi Yang and
                  Jingren Zhou and
                  Junyang Lin and
                  Kai Dang and
                  Keming Lu and
                  Keqin Bao and
                  Kexin Yang and
                  Le Yu and
                  Mei Li and
                  Mingfeng Xue and
                  Pei Zhang and
                  Qin Zhu and
                  Rui Men and
                  Runji Lin and
                  Tianhao Li and
                  Tingyu Xia and
                  Xingzhang Ren and
                  Xuancheng Ren and
                  Yang Fan and
                  Yang Su and
                  Yichang Zhang and
                  Yu Wan and
                  Yuqiong Liu and
                  Zeyu Cui and
                  Zhenru Zhang and
                  Zihan Qiu},
  title        = {Qwen2.5 Technical Report},
  journal      = {CoRR},
  volume       = {abs/2412.15115},
  year         = {2024},
  url          = {https://doi.org/10.48550/arXiv.2412.15115},
  doi          = {10.48550/ARXIV.2412.15115},
  eprinttype   = {arXiv},
  eprint       = {2412.15115},
  bibsource    = {dblp computer science bibliography, https://dblp.org}
}

@misc{beltagy2020longformerlongdocumenttransformer,
      title={Longformer: The Long-Document Transformer}, 
      author={Iz Beltagy and Matthew E. Peters and Arman Cohan},
      year={2020},
      eprint={2004.05150},
      archivePrefix={arXiv},
      primaryClass={cs.CL},
      url={https://arxiv.org/abs/2004.05150}, 
}

@misc{liu2019robertarobustlyoptimizedbert,
      title={RoBERTa: A Robustly Optimized BERT Pretraining Approach}, 
      author={Yinhan Liu and Myle Ott and Naman Goyal and Jingfei Du and Mandar Joshi and Danqi Chen and Omer Levy and Mike Lewis and Luke Zettlemoyer and Veselin Stoyanov},
      year={2019},
      eprint={1907.11692},
      archivePrefix={arXiv},
      primaryClass={cs.CL},
      url={https://arxiv.org/abs/1907.11692}, 
}

@article{DBLP:journals/corr/abs-2106-09685,
  author       = {Edward J. Hu and
                  Yelong Shen and
                  Phillip Wallis and
                  Zeyuan Allen{-}Zhu and
                  Yuanzhi Li and
                  Shean Wang and
                  Weizhu Chen},
  title        = {LoRA: Low-Rank Adaptation of Large Language Models},
  journal      = {CoRR},
  volume       = {abs/2106.09685},
  year         = {2021},
  url          = {https://arxiv.org/abs/2106.09685},
  eprinttype   = {arXiv},
  eprint       = {2106.09685},
  bibsource    = {dblp computer science bibliography, https://dblp.org}
}

@article{DBLP:journals/corr/abs-1908-09203,
  author       = {Irene Solaiman and
                  Miles Brundage and
                  Jack Clark and
                  Amanda Askell and
                  Ariel Herbert{-}Voss and
                  Jeff Wu and
                  Alec Radford and
                  Jasmine Wang},
  title        = {Release Strategies and the Social Impacts of Language Models},
  journal      = {CoRR},
  volume       = {abs/1908.09203},
  year         = {2019},
  url          = {http://arxiv.org/abs/1908.09203},
  eprinttype   = {arXiv},
  eprint       = {1908.09203},
  bibsource    = {dblp computer science bibliography, https://dblp.org}
  }
\bibliographystyle{acl_natbib}

\clearpage
\appendix

\section{\texorpdfstring{ {Dataset Details}}{Dataset Details}}
\label{sec:appendix-data}

 {This section complements Section~3 with the split statistics of \dataname, the request pool used for \fullyllmreview, and an audit of the surface cues removed by test debiasing.}

\paragraph{ {Split statistics.}}
 {Table~\ref{tab:dataset_stats} gives the number of sentences per provenance label, generation condition, and split for each generator.}

\begin{table*}[t]
\centering
\setlength{\tabcolsep}{3pt}
\begin{adjustbox}{max width=\textwidth}
\begin{tabular}{l rrr rrr rrr}
\toprule
& \multicolumn{3}{c}{\textit{Train-raw}} & \multicolumn{3}{c}{\textit{Dev}} & \multicolumn{3}{c}{\textit{Test-raw}}\\
\cmidrule(lr){2-4}\cmidrule(lr){5-7}\cmidrule(lr){8-10}
\textbf{Condition} & Hum. & Hyb. & AI & Hum. & Hyb. & AI & Hum. & Hyb. & AI\\
\midrule
\multicolumn{10}{l}{\textbf{GPT-4o}}\\
\quad\fullyhumanreview & 61,993 & -- & -- & 20,500 & -- & -- & 20,500 & -- & --\\
\quad\llmparaphrase & 48,816 & 1,127 & 57 & 20,015 & 462 & 23 & 20,015 & 462 & 23\\
\quad\llmenrichbullets & 13,177 & 34,761 & 2,062 & 5,403 & 14,252 & 845 & 5,376 & 14,182 & 842\\
\quad\fullyllmreview & -- & -- & 50,000 & -- & -- & 20,500 & -- & -- & 20,500\\
\cmidrule(l){1-10}
\quad\textit{Total} & 123,986 & 35,888 & 52,119 & 45,918 & 14,714 & 21,368 & 45,891 & 14,644 & 21,365\\
\quad\textit{All} & \multicolumn{3}{c}{\textbf{211,993}} & \multicolumn{3}{c}{\textbf{82,000}} & \multicolumn{3}{c}{\textbf{81,900}}\\
\midrule
\multicolumn{10}{l}{\textbf{Gemini-3.5-Flash-Lite}}\\
\quad\fullyhumanreview & 58,622 & -- & -- & 20,500 & -- & -- & 20,500 & -- & --\\
\quad\llmparaphrase & 46,961 & 2,925 & 114 & 19,254 & 1,199 & 47 & 19,254 & 1,199 & 47\\
\quad\llmenrichbullets & 11,661 & 36,948 & 1,391 & 1,101 & 3,490 & 131 & -- & -- & --\\
\quad\fullyllmreview & -- & -- & 50,000 & -- & -- & 20,500 & -- & -- & 20,500\\
\cmidrule(l){1-10}
\quad\textit{Total} & 117,244 & 39,873 & 51,505 & 40,855 & 4,689 & 20,678 & 39,754 & 1,199 & 20,547\\
\quad\textit{All} & \multicolumn{3}{c}{\textbf{208,622}} & \multicolumn{3}{c}{\textbf{66,222}} & \multicolumn{3}{c}{\textbf{61,500}}\\
\midrule
\multicolumn{10}{l}{\textbf{DeepSeek-V3.1}}\\
\quad\fullyhumanreview & 49,125 & -- & -- & 9,900 & -- & -- & 9,900 & -- & --\\
\quad\llmparaphrase & 45,941 & 3,948 & 111 & 9,096 & 782 & 22 & 9,096 & 782 & 22\\
\quad\llmenrichbullets & 3,184 & 44,659 & 2,157 & 630 & 8,843 & 427 & 603 & 8,458 & 409\\
\quad\fullyllmreview & -- & -- & 50,000 & -- & -- & 9,900 & -- & -- & 9,900\\
\cmidrule(l){1-10}
\quad\textit{Total} & 98,250 & 48,607 & 52,268 & 19,626 & 9,625 & 10,349 & 19,599 & 9,240 & 10,331\\
\quad\textit{All} & \multicolumn{3}{c}{\textbf{199,125}} & \multicolumn{3}{c}{\textbf{39,600}} & \multicolumn{3}{c}{\textbf{39,170}}\\
\bottomrule
\end{tabular}
\end{adjustbox}
\caption{Final corpus statistics after splitting and normalization.}
\label{tab:dataset_stats}
\end{table*}

\paragraph{\fullyllmreview~request pool.}
Table~\ref{tab:prompt-pool} lists the review requests sampled for \fullyllmreview.

\begin{table*}[t]
\footnotesize
\setlength{\tabcolsep}{4pt}
\renewcommand{\arraystretch}{1.05}
\begin{tabular}{@{}r p{0.93\textwidth}@{}}
\toprule
& \textbf{Review Prompts} \\
\midrule
\multicolumn{2}{@{}l}{\textit{(a) Minimal, unstructured requests}}\\
1 & Review this paper \\
2 & Write a review of this paper \\
3 & Can you review the attached paper? \\
4 & Provide a review for this manuscript \\
5 & What would a review of this paper look like? \\
6 & How would you review this paper? \\
7 & Could you evaluate this manuscript? \\
\addlinespace[2pt]

\multicolumn{2}{@{}l}{\textit{(b) Requests specifying depth or tone}}\\
8 & Write a critical review of this paper \\
9 & Give me a detailed peer review of this manuscript \\
10 & Provide a constructive review for this paper \\
11 & Write an academic review for this research paper \\
12 & Give me a thorough, professional review \\
13 & Write a thorough review with line-by-line comments \\
\addlinespace[2pt]

\multicolumn{2}{@{}l}{\textit{(c) Role- and venue-framed requests}}\\
14 & Act as a peer reviewer and review this paper \\
15 & Pretend you're reviewing this for a top-tier conference \\
16 & Review this paper for submission to a top venue \\
17 & Help me review this paper I'm assigned to peer review \\
18 & Act as an anonymous reviewer for a top-tier ML conference and write a full review \\
19 & Pretend you're a skeptical reviewer and write a critical review \\
20 & As an expert in this area, review this paper for correctness and novelty \\
\addlinespace[2pt]

\multicolumn{2}{@{}l}{\textit{(d) Requests specifying output structure}}\\
21 & Review this paper and include strengths and weaknesses \\
22 & Write a review covering methodology, results, and conclusions \\
23 & Review this paper with sections on novelty, technical quality, and clarity \\
24 & Provide a peer review with summary, major comments, and minor comments \\
25 & Please provide a NeurIPS-style review with scores and justification \\
26 & Draft an OpenReview-style review (summary, strengths, weaknesses, questions, rating) \\
27 & Write a review with: Summary, Strengths, Weaknesses, Questions for Authors, and Overall Recommendation \\
28 & Provide (1) brief summary, (2) novelty/impact, (3) soundness, (4) clarity, (5) reproducibility, (6) detailed comments \\
29 & Give a review in bullet points with actionable suggestions \\
30 & Write two sections: `Major Concerns' and `Minor Comments' \\
\addlinespace[2pt]

\multicolumn{2}{@{}l}{\textit{(e) Requests specifying evaluation criteria or a decision}}\\
31 & Write a review recommending accept/reject for this paper \\
32 & Review this paper and suggest a decision \\
33 & Evaluate this paper on novelty, technical quality, clarity, and significance---then give an accept/reject recommendation \\
34 & Assess methodological rigor (baselines, ablations, stats, datasets) and list missing experiments \\
\bottomrule
\end{tabular}
\caption{The 34 review-request prompts sampled for the \fullyllmreview~condition. One
prompt is drawn uniformly at random per generated review. Prompts vary in specificity,
role framing, required output structure, and evaluative stance, so that the condition is
not tied to a single prompting style. Group sizes: (a)~7, (b)~6, (c)~7, (d)~10, (e)~4.}
\label{tab:prompt-pool}
\end{table*}

\paragraph{ {Debiasing audit.}}
 {Table~\ref{tab:debias-audit} reports how often each surface marker occurs before and after debiasing. Style neutralization (v1) nearly eliminates modal hedges, formal transitions, emphatic adjectives, first-person pronouns, and abbreviations, while content canonicalization (v2) further reduces suggestion language, negative-critique terms, numbered references, and sentence length.}

\begin{table}[t]
\centering\small
\setlength{\tabcolsep}{4pt}
\begin{tabular}{@{}lrrr@{}}
\toprule
\textbf{Surface marker} & \textbf{Raw} & \textbf{v1} & \textbf{v2} \\
\midrule
Modal hedges            & 23.9 & \phantom{0}0.5 & \phantom{0}0.7 \\
Passive voice           & 13.9 & \phantom{0}3.2 & \phantom{0}8.9 \\
Formal transitions      & 10.9 & \phantom{0}0.0 & \phantom{0}0.0 \\
Emphatic adjectives     & \phantom{0}6.4 & \phantom{0}0.1 & \phantom{0}0.3 \\
First-person pronouns   & \phantom{0}7.9 & \phantom{0}0.2 & \phantom{0}0.2 \\
Abbreviations           & \phantom{0}3.7 & \phantom{0}0.0 & \phantom{0}0.0 \\
Suggestion language     & --- & 21.0 & 16.5 \\
Negative-critique terms & --- & 13.0 & \phantom{0}7.7 \\
Numbered refs.          & --- & \phantom{0}4.7 & \phantom{0}0.7 \\
Comparison framing      & --- & \phantom{0}4.8 & \phantom{0}3.3 \\
\midrule
Mean length (words)     & 22.5 & 21.7 & 18.7 \\
\bottomrule
\end{tabular}
\caption{Surface-marker incidence (\% of sentences) on the GPT test split across the
three normalisation levels (Raw: Test-raw; v1/v2: Debiased-Test-v1/v2; numbered refs.\ = Sec./Tab./Fig.). Dashes mark patterns not instrumented at the original
level.}
\label{tab:debias-audit}
\end{table}

\clearpage
\section{\texorpdfstring{{Prompts}}{Prompts}}
\label{sec:appendix-prompts}

{We list the prompts that determine the provenance labels, the debiased test sets, the analysis scores, and the in-context baselines, since these directly affect how our results should be read. Placeholders are shown in braces, long rule lists are abridged with [\dots], and the complete prompts are released with the code. The generation prompts for \llmparaphrase~and \fullyllmreview~are described in Section~3 and Table~\ref{tab:prompt-pool}.}

\subsection{\texorpdfstring{{Bullet extraction for \llmenrichbullets}}{Bullet extraction for LLM-Expand}}
\label{app:prompt-extract}

{Step~(i) of \llmenrichbullets~is designed as a compression of the human review rather than a second generation step: it runs at temperature $0.1$, forbids adding content, and its output is further filtered by exact match against the review (Section~3). This is what licenses the \labelhumaninsight~label for ideas that survive into the expanded review.}

\begin{lstlisting}
Extract the key points from the following review as short bullet points.
1. ONLY extract 3-5 points that are EXPLICITLY STATED in the original review text
2. Do NOT add ANY information, opinions, or insights not present in the original review
3. Do NOT interpret, infer, or expand beyond what is written
[...]
6. Use the EXACT terminology and phrasing from the review as much as possible
7. Cover ONLY weaknesses

Review text: """{REVIEW_TEXT}"""
\end{lstlisting}

\subsection{\texorpdfstring{{Provenance annotation}}{Provenance annotation}}
\label{app:prompt-annot}

{The judge runs at temperature $0.1$ and is asked an entailment question about idea content, not a stylistic one; it never sees which generator produced the sentence. Replies that do not match the required format after three attempts are dropped, which is why the rows of Table~\ref{tab:source-label} do not sum to 100\%.}

\begin{lstlisting}
You will be given:
  1. An original HUMAN review of an academic paper
  2. A single SENTENCE from an LLM-generated review based on the human review
Your task: Determine whether the core idea/insight in the SENTENCE originates from
the HUMAN review or was newly introduced by the AI.

=== ORIGINAL HUMAN REVIEW ===
{HUMAN_REVIEW}
=== SENTENCE TO CLASSIFY ===
{SENTENCE}

Assuming you are a system for textual entailment or natural language inference:
  * If this sentence can be FULLY inferred/entailed from the human review -> "Human"
  * If the meaning/idea/insights PARTIALLY inferred/entailed from the human review -> "Hybrid"
  * If NOTHING (the meaning/idea/insights) comes from the human review -> "AI"
Respond in EXACTLY this format (nothing else):
Label: Human OR AI OR Hybrid
Explanation: [one brief sentence explaining your reasoning]
\end{lstlisting}

\subsection{\texorpdfstring{{Test debiasing}}{Test debiasing}}
\label{app:prompt-debias}

 {Both debiased sets are rewritten sentence by sentence at temperature $0.2$. Labels and paper assignments are carried over unchanged, so each debiased set is a paired rewrite of $Test$-$raw$ rather than a new sample. For Debiased-Test-v1, a second pass is applied only to sentences that still match one of the rule patterns, so already-clean sentences are not paraphrased twice. Debiased-Test-v2 is applied to the v1 output.}

\begin{lstlisting}[title={Debiased-Test-v1 (style neutralization)}]
Rewrite the peer-review sentence so the critique CONTENT is fully preserved
but every stylistic signal that reveals authorship (human vs. AI) is removed.
1. NO first-person    2. NO abbreviations    3. NO formal openers
4. NO emphatic adjectives    5. NO passive voice    6. NO modal hedges
7. NORMALIZE length - aim 12-25 words    8. NO informal punctuation
[...]
\end{lstlisting}

\begin{lstlisting}[title={Debiased-Test-v2 (content canonicalization)}]
Identify the CORE IDEA or OBSERVATION in the sentence and express it in a
completely neutral, standardized form so that a reader can only judge the
intellectual content, not who wrote it.
1. DECLARATIVE FACTS ONLY - no suggestions, no recommendations, no imperatives.
     BAD : "The authors should add ablations."
     GOOD: "Ablation experiments are absent from the paper."
2. NO AUTHORSHIP VOICE    3. NO EVALUATIVE STANCE    4. NO MODAL HEDGES
5. NO PASSIVE VOICE    6. NO FORMAL TRANSITIONS    7. NO NUMBERED REFERENCES
8. NO COMPARISON FRAMING    9. STRICT LENGTH: 12-18 words per sentence
[...]
\end{lstlisting}

\subsection{\texorpdfstring{ {In-context baselines}}{In-context baselines}}
\label{app:prompt-icl}

 {To give the prompted baselines a fair chance on the debiased sets, the system prompt explicitly asks for idea provenance rather than style and warns that some sentences have been style-normalized. Sentences are classified in batches of 20 at temperature $0$; the ten demonstrations come from the training split only.}

\begin{lstlisting}[title={System message (abridged)}]
Classify the PROVENANCE OF THE IDEA expressed in each peer-review sentence -
i.e. where the intellectual contribution came from - into exactly one of three
classes. You are NOT judging writing style, fluency, or grammar.
1) human   The insight originates from a human expert reviewer's own reading
           and judgement of the paper. [...]
2) hybrid  The sentence comes from a review whose content is genuinely mixed -
           some points are AI-generated and some are human-generated. [...]
3) AI      The idea was produced by an AI system with no meaningful human
           intellectual contribution. [...]
CRITICAL RULES
- Judge IDEA PROVENANCE, not surface style. Some of these sentences have been
  deliberately style-normalised to strip authorship cues - in those cases you
  must reason from the SUBSTANCE of the point alone.
- "hybrid" is a real content class, not an "unsure" bucket.
[...]
\end{lstlisting}

\end{document}